# "Predictive Modelling of Air Quality Index (AQI) Across Diverse Cities and States of India using Machine Learning: Investigating the Influence of Punjab's Stubble Burning on AQI Variability"

Kamaljeet Kaur Sidhu[1], Dr. Habeeb Balogun[1] and Dr. Kazeem Oluwakemi Oseni[2]

[1] School of Computer Science and Engineering, University of Westminster, United Kingdom
[2] School of Computer Science and Technology, University of Bedfordshire, United Kingdom

## Abstract

*Air pollution is a common and serious problem nowadays and it cannot be ignored as it has harmful impacts on human health. To address this issue proactively, people should be aware of their surroundings, which means the environment where they survive. With this motive, this research has predicted the AQI based on different air pollutant concentrations in the atmosphere.*

*The dataset used for this research has been taken from the official website of CPCB. The dataset has the air pollutant concentration from 22 different monitoring stations in different cities of Delhi, Haryana, and Punjab. This data is checked for null values and outliers. But, the most important thing to note is the correct understanding and imputation of such values rather than ignoring or doing wrong imputation. The time series data has been used in this research which is tested for stationarity using The Dickey-Fuller test.*

*Further different ML models like CatBoost, XGBoost, Random Forest, SVM regressor, time series model SARIMAX, and deep learning model LSTM have been used to predict AQI. For the performance evaluation of different models, I used MSE, RMSE, MAE, and R2. It is observed that Random Forest performed better as compared to other models.*

## 1. Introduction

Putting all the life threats aside, air pollution is the deadliest health threat to all species. In other words, it can be called a silent killer. It is estimated by WHO, that around 7 million people die every year this is because of the presence of deadly fine particles in the air, which lead to diseases such as stroke, heart disease, lung cancer, chronic obstructive pulmonary diseases, and respiratory infections, including pneumonia. In the list of most polluted countries, India specifically the capital city Delhi is in the top ten where the air quality reaches hazardous levels.

Cancellation of flight is a common issue faced by people at Delhi International Airport due to reduced visibility associated with air pollution [1].

The Air Quality Index (AQI) is a daily basic expected measure of air pollution levels in the environment. It informs about the short-term potential health effects on well-being. Being aware of AQI helps to safeguard everyone from the effects of bad air quality. Even low levels of AQI can impact the health of susceptible individuals like old people, children, and patients.

According to the Indian Government (CPCB), the AQI ranges from 0-500, with 0 being good and 500 being severe. There are eight major pollutants to be considered for AQI calculation, viz: PM 10 and PM 2.5, CO, $O_3$, $NO_2$, $SO_2$, NH3, and Pb. To calculate AQI, data for a minimum of three pollutants is required, of which one should be either PM10 or PM2.5. Each air pollutant hasdifferent health effects. However, the data availability, averaging period, monitoring frequency, and measurement methods highly impact the selection of air pollutants to calculate AQI [2]. Initially, for this study daily concentrations of PM2.5 (ug/m3), PM10 (ug/m3), NO (ug/m3), NO2 (ug/m3), NOx (ppb), NH3 (ug/m3), SO2 (ug/m3), CO (mg/m3), Ozone (ug/m3), Benzene (ug/m3), Toluene (ug/m3), Temp (degree C), RH (%), WS (m/s), WD (degree), SR (W/mt2), Xylene (ug/m3) from 1st January 2019 to mid-October 2023 is considered.

Hence, ML is one of the techniques that can be used to predict or suggest a solution for air pollutants. The collection of raw data, which is of not much use, once clubbed with various machine learning algorithms can be used to accomplish the task for example, predict the future. For this research, different ML regression algorithms are considered including CatBoost, XGBoost, Random Forest, and SVR, also, from a set of time series forecasting algorithms SARIMAX and from deep learning models LSTM is used to predict AQI for different monitoring stations in Delhi, Haryana, and Punjab.

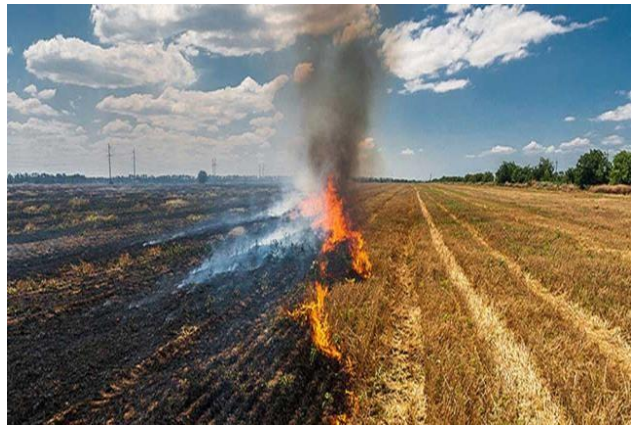

Figure 1. Stubble burning

This research is focused on answering the following questions.

1. How machine learning can be used to predict AQI?
2. What are the spatial and temporal patterns of AQI across different states?
3. Which is the most effective ML algorithm and has the most accuracy in predicting the AQI?

## 2. LITERATURE REVIEW

**Models**

[3] They collected the data from CPCB from July 2017 to September 2022. They determined particulate matter, gaseous pollutants, and metrological factors for the prediction of AQI in Visakhapatnam, Andhra Pradesh, India. Among all, they considered $PM_{2.5}$, $PM_{10}$, $CO_2$, CO, SOX, $NO_X$, $O_3$, and $NH_3$ as key contributors to AQI based on the correlation matrix. However, they found that the contribution of meteorological factors like Temperature, Relative Humidity, Wind Speed, Wind Direction, Solar Radiation, Air Pressure, Ambient Temperature, Rainfall and Total Rainfall in AQI prediction is almost negligible. To identify the best parameters for ML models they used Grid Search techniques and performed hyperparameter tuning on models to achieve the best performance. This research paper concludes that the Random Forest and CatBoost algorithms outperformed other machine learning models like LightGBM, Adaboost, and XGboost in predicting AQI accurately.

[4] This research aims to compare the various models such as SARIMA, SVM, and LSTM for the prediction of AQI for Ahmedabad city of Gujarat, India. LSTM has been used in [5] [6] [7], and [8] for the same task. LSTM has shown a quick convergence and reduced training cycle with good accuracy [9]. The performance of LSTM and ARIMA is compared in [33], where LSTM is found to be far superior to ARIMA by producing the least mean absolute error (MAE) and RMSE values.

Research conducted by [4] presents a comparative analysis of SVM, LSTM, and SARIMA ML models. They found the SVM with radial basis function (RBF) kernel provides better results for AQI prediction. They have not considered the records for Benzene, Toluene, and Xylene (BTX) due to null values and noisy information. They used Z-Score for outlier processing. But there is another interactive way to do it by creating visualization in Tableau.

Deleting the rows with missing values is not a good approach as it can lead to information loss,which would result in wrong output [10]. However, in [4] a season-specific mean value is used to replace the null value observed in that season only. Whereas, instead of season-specific, temporal, and spatial (location) specific mean values can be used to fill in missing values.

[11] To check if the data is stationary or not the Dickey-Fuller Test has been used. They used the p-value to check the dataset, i.e. it shows whether the data is stationary or not. They noticed seasonality and trends in the data. So, they used SARIMA and Holt-Winter's for the prediction of AQI. They used MAPE as the score function to analyze the performance of models. For performance evaluation of both models is done by calculating and comparing the MAPE. They observed the SARIMA model outperforms in comparison to the Holt-Winter's model.

[4] In this research, they used SARIMA and LSTM to predict the AQI for just one city named Ahmedabad city of Gujarat, India. However, I intend to predict multiple monitoring stations for Delhi, Haryana, and Punjab. In [12] paper Google's Street View data and ML models have been used to predict air quality at different places in Oakland City, California. He developed a web application to predict air quality for any location in the city's neighbourhoods.

[13] They revealed that meteorological variables help the city reduce air pollution naturally, especially during the summer. However, air pollution becomes a significant issue over the winter due to the city's low height and seasonal temperature inversion. [14] They revealed that $PM_{2.5}$, $PM_{10}$, $CO_2$, CO, SOX, NOX, $O_3$, and $NH_3$ are key contributors to AQI. Thus, an increased AQI

leads to various harmful environmental conditions like global warming, acid rain, the development of smog and aerosols, decreased visibility, climate change etcetera.

The authors of [15] have stated, the regression ML algorithms are efficient in determining the AQI. They considered statistical criteria like MAE, MAPE, R, and RMSE for evaluating the performance of regression models. In this study, some of those statistical metrics are considered for evaluating the model's performance.

## 2.1. Influence of Stubble Burning on AQI

In a research paper [16], they revealed the major contributor to air pollution especially in South Asia is stubble burning. It is a significant source of gaseous pollutants causing serious damage to human health and the environment. This paper highlights the seriousness of this issue by revealing its findings. It was reported that the burning of 63 Mt of crop stubble releases 3.4 Mt of CO, 0.1 Mt of NOx, 91 Mt of $CO_2$, 0.6 Mt of $CH_4$, and 1.2 Mt of PM into the atmosphere.

The intensive rice-wheat rotation system which generates a large amount of stubble is the main reason for bad air quality in India. Each year about 352 Mt of stubble is generated in India straight after the harvesting season. Out of which 22% and 34% are contributed by wheat and rice stubble respectively. Thus, about 84 Mt (23.86%) of the stubble is burnt on-field each year. It is common to observe the disastrous haze over India during the winter because of stubble burning as it coincides with the burning periods from October to December.

[17] They employed the Moderate Resolution Imaging Spectro radiometer active fire products and TROPOspheric Monitoring Instrument products on CO and $NO_2$ concentrations for spatio-temporal assessment of stubble burning and associated emissions. They used Google Earth Engine (GEE) to perform analysis. They found about three times rise in crop residue burning in November than in October, with 92.58% and 7.42% reported from Punjab and the Haryana states in November.

Since 1990, Delhi has been one of the most polluted cities in the world. In a research paper [18] they found that in 2019, the global air quality report showed that 14 out of the 20 most polluted cities in the world are from India. [19] They highlighted the burning of firecrackers during the Diwali festival as the cause of poor winter air in India. However, [18] [20] mentioned, that it is not the only reason for air quality problems during the winter because the presence of harsh pollution begins even before the festival time. However, [20] reported the burning of firecrackers results in a quantitatively small and statistically significant rise in air pollution. They compared the stubble-burning periods with and without Diwali and found an increase of approximately 40 µg/m3 PM2.5 concentration during Diwali days in 2018.

Further [20] mentioned, that during the rice stubble burning season the impact of stubble burning is more severe as the lower winter temperature leads to a more stable atmosphere (Inversion conditions). There is a fact that pollutants stay longer in the atmosphere during this time and that the amount of rice stubble burned is quite higher than that of wheat. Therefore, a harsh level of pollution often obstructs visibility. In Delhi, the air pollution level during October 2017 was six-fold as compared to that during July of the same year. The atmospheric inversion provides a greater life span for pollutants, poorer dispersion, and a lesser rate of smoke diffusion.

## 3. DATA UNDERSTANDING

### Dataset Description

The dataset used for this research is downloaded from the official website of CPCB [32].https://airquality.cpcb.gov.in/ccr/#/caaqm-dashboard-all/caaqm-landing/data it is around five years of data from 1st January 2019 to mid-October 2023. The raw data has a total of 21 columns. As it is a daily basics data the temporal information about each record is stored in Date. For maintaining the information about the locations there are three Columns State, City, and Monitoring Station. Overall, the collected data is for three states Delhi, Punjab, and Haryana. Further, the data about 15 different monitoring stations in 15 different cities is collected from Haryana. However, from Punjab, data about six monitoring stations in six different cities is collected. Further, there are different air pollutants recorded for each day like PM2.5 (ug/m3), PM10 (ug/m3), NO (ug/m3), NO2 (ug/m3), NOx (ppb), NH3 (ug/m3), SO2 (ug/m3), CO (mg/m3) and Ozone (ug/m3). There are three Volatile Organic Compounds Benzene (ug/m3), Toluene (ug/m3), and Xylene (ug/m3). Moreover, there are four meteorological factors Temp (degree C), RH (%), WS (m/s), WD (degree), and SR (W/mt2). The measurement unit for all these factors is mentioned with the column names.

## 4. DATA PREPROCESSING

Data pre-processing is the process of data understanding, identification, and specification of data-related issues as well as a knowledge-based approach that can be utilized to address these issues. Thus, it enhances data reliability for use in ML studies [30]. The two most common errors in data are missing values and outliers.

### Missing Values

Missing value means the value or data that is not stored for variables in a dataset. In pandas missing values are represented by NaN. In this dataset, the total records are 38277. The number of missing values for different columns is shown in Figure 5.1 along with the total percentage of missing values for different variables.

Your selected dataframe has 21 columns.
There are 17 columns that have missing values.

| | Missing Values | % of Total Values |
|---|---|---|
| Temp (degree C) | 31746 | 82.900000 |
| Xylene (ug/m3) | 18047 | 47.100000 |
| Toluene (ug/m3) | 8944 | 23.400000 |
| WD (deg) | 8844 | 23.100000 |
| RH (%) | 8178 | 21.400000 |
| SR (W/mt2) | 7885 | 20.600000 |
| Benzene (ug/m3) | 6216 | 16.200000 |
| WS (m/s) | 5377 | 14.000000 |
| NH3 (ug/m3) | 4277 | 11.200000 |
| Ozone (ug/m3) | 4047 | 10.600000 |
| NO (ug/m3) | 1188 | 3.100000 |
| NOx (ppb) | 1092 | 2.900000 |
| PM2.5 (ug/m3) | 1066 | 2.800000 |
| SO2 (ug/m3) | 917 | 2.400000 |
| PM10 (ug/m3) | 792 | 2.100000 |
| CO (mg/m3) | 707 | 1.800000 |
| NO2 (ug/m3) | 687 | 1.800000 |

Figure 2. Missing Values

Firstly, missing values are imputed with mean where all the records are aggregated based on Monitoring station, year, and month. The mean of monthly records for each year is calculated as per the monitoring stations. Then the missing values are filled with the calculated mean for that specific year and month as per monitoring stations. For example, for monitoring station Golden Temple, Amritsar – PPCB the mean is calculated for January 2019. Then whatever the missing values are there for January 2019 for Golden Temple, Amritsar – PPCB are filled with this mean.

Even after mean imputation some missing values are found in the data. So, the data frame is converted into a SQL table. Then, a query is generated to calculate the number of NaN values for each column for every month of each year as per monitoring stations. After analysis, it is found that for some stations there are no records for the whole month. Further ensuring the availability of data for at least one of the pollutant matters (PM2.5 and PM10) and at least three out of the seven pollutants to find AQI the remaining missing values are filled with 0.

### Correlation Matrix

The Pearson correlation coefficient is used to summarize the strength of the linear relationship between the numerical variables in the dataset.

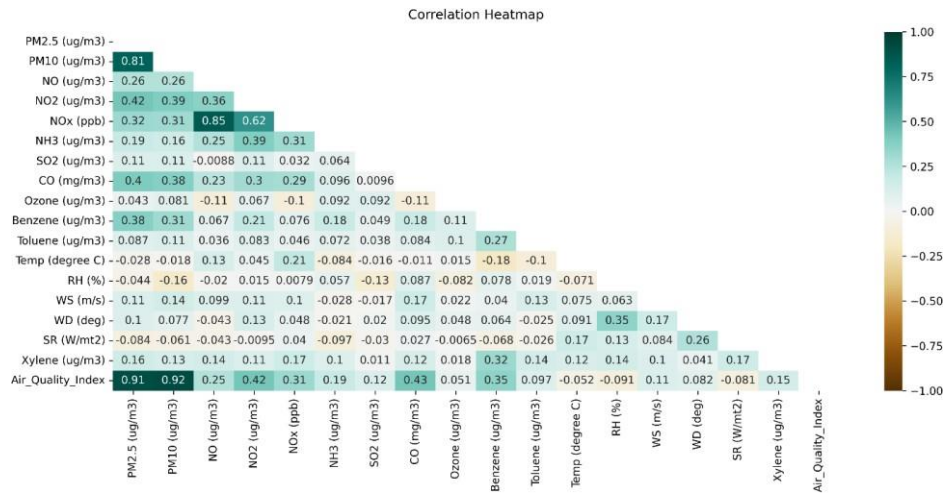

Figure 3. Correlation Matrix

It is analysed that both the pollutant matter PM 2.5 and PM 10 have the highest positive correlation with AQI. However, the meteorological factors are least correlated, in fact, Temperature, SR, and RH are negatively correlated with AQI.

### Feature Selection

For this research it is not a good approach to consider variables with weak correlation, also there is not sufficient and authentic information available about the concentration of these variables. So, the meteorological factors, Volatile Organic Compounds, and Ozone are removed from the dataset.

**Statistical Description**

|       | PM2.5 (ug/m3) | PM10 (ug/m3) | NO (ug/m3) | NO2 (ug/m3) | NOx (ppb) | NH3 (ug/m3) | SO2 (ug/m3) | CO (mg/m3) | Air_Quality_Index |
|-------|---------------|--------------|------------|-------------|-----------|-------------|-------------|------------|-------------------|
| count | 38277.000000  | 38277.000000 | 38277.000000 | 38277.000000 | 38277.000000 | 38277.000000 | 38277.000000 | 38277.000000 | 38277.000000 |
| mean  | 62.147408     | 132.648127   | 13.346512  | 22.942600   | 26.383201 | 31.765125   | 12.029370   | 0.846457   | 142.614213        |
| std   | 48.286427     | 83.621161    | 18.614464  | 16.881163   | 22.803996 | 31.369546   | 10.475631   | 0.599170   | 92.931801         |
| min   | 0.000000      | 0.000000     | 0.000000   | 0.010000    | 0.000000  | 0.000000    | 0.000000    | 0.000000   | 1.600000          |
| 25%   | 30.620000     | 72.700000    | 4.540000   | 10.880000   | 12.910000 | 11.660000   | 5.820000    | 0.460000   | 77.200000         |
| 50%   | 49.340000     | 113.890000   | 8.610000   | 19.110000   | 21.130000 | 25.000000   | 9.250000    | 0.690000   | 113.953333        |
| 75%   | 79.600000     | 171.650000   | 14.900000  | 29.590000   | 32.560000 | 42.310000   | 15.100000   | 1.050000   | 180.033333        |
| max   | 999.990000    | 999.990000   | 467.630000 | 259.570000  | 471.790000 | 498.400000  | 187.600000  | 7.530000   | 1112.487500       |

Figure 4. Descriptive Statistics of the Dataset

It is analysed, that there is an extreme gap not only between the mean and the maximum value but also in the 75% percentile for each variable. It might be due to the existence of outliers in the data. This statistical description showcases the need for further preprocessing.

**Outliers**

Outliers mean the data points that deviate significantly from the rest of the observations, indicating either exceptionally good or poor air quality. It is a critical step to identify and understand the extreme pollution events that can lead to serious environmental and public health issues. Outliers can be caused by a variety of variables, including measurement mistakes, odd meteorological circumstances, and pollution occurrences.

For identifying the outliers in data, I used Tableau and displayed it by Box Plots. A box plot helps to visualize the distribution of data for a continuous variable. It helps to visualize the centre and spread of data on a scale. The extended lines from the box are called whiskers. The whiskers represent the expected variation of the data. The whiskers are 1.5 times the IQR extension from the top and bottom of the box. If the data do not extend to the end of the whiskers, then the whiskers extend to the minimum and maximum data values. An outlier is more extreme than the expected variation. Further, these data points need to be reviewed to determine if they are outliers or errors; the whiskers will not include these outliers.

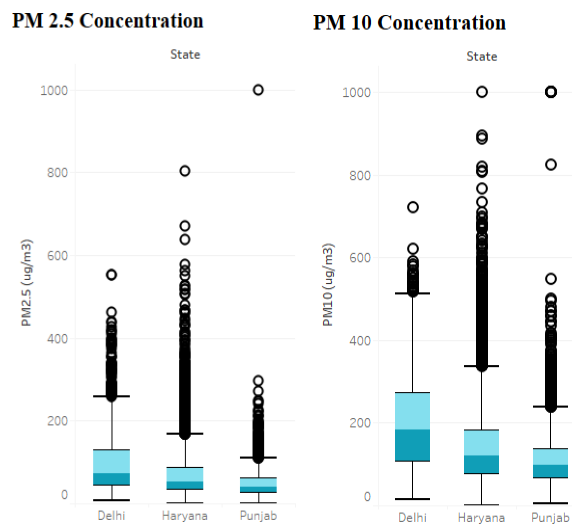

Figure 5. Outliers in PM 2.5 and PM 10 Concentration

International Journal of Managing Information Technology (IJMIT) Vol.16, No.1, February 2024

This process is repeated for all the air pollutants to identify the outliers. But on the basics of domain knowledge and data understanding, all the dots cannot be considered outliers. In experimentation, it is analysed that an attempt to remove all the outliers leads to entire data loss. Thus, considering all the dots as outliers and deleting them is not a good approach. A simple definition of an outlier is "data that does not belong to other data points". In this study, the same patterns are noticed, where the data points appear far from the other data present. Even during the investigation, some scenarios are found to be unusual as compared to the rest of the data.

### Feature Engineering Techniques for ML

#### One-Hot Encoding

In this research binary encoding scheme is used, which converts the columns into 0 or 1. A specific variable that is true for a record will be marked as 1, however, the rest of the variables of the same category will be marked 0. Overall, this process aims to assign a unique value for each possible case.

In this research one-hot encoding is applied to spatial variables like state and city. Because variables having data type 'Object' cannot be included in ML algorithms. However, another variable Monitoring Station has not been considered due to the high volume of values. Because adding too many columns in a dataset impacts the model's performance while training and testing a large dataset. But excluding the Monitoring Station column will not result in a data loss, because each city has only a Monitoring station. So, including the city will be enough to make a record unique.

#### Scaling

The process of normalizing the range of features in a dataset refers to feature scaling. The features of the dataset used in this research vary per unit. Therefore, feature engineering is implemented for ML models to interpret these features on the same scale. In this research Standard scaling is used for this task.

## 5. STUDY AREA

This research focuses on three major states Delhi, Punjab, and Haryana, which are the neighbouring states.

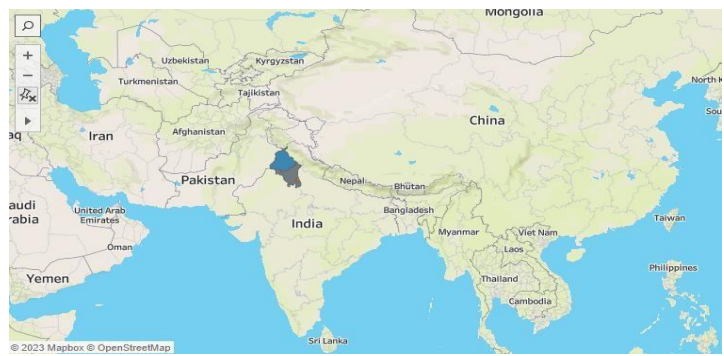

Figure 6. Research Area on Map
22

This research is focused on states in the Northern part of India. Further, the different monitoring stations included in this research are shown revealing the concentration of AQI during the last peak period, November 2022.

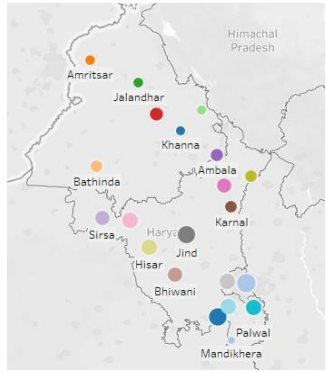

Figure 7 Different cities where Monitoring Stations are located on the Map.

In total, the dataset from 22 different monitoring stations has been used. However, due to the limitations of Tableau names are not appearing for every state.

In different papers [16] [17] they have mentioned the stubble-burning activities in Punjab and Haryana and its impact on other neighbouring countries, especially Delhi. During the peak period mainly November, it is believed that stubble burning, and smoke generated in Punjab and Haryana are responsible for the bad air quality of Delhi. The EDA conducted in this research focuses on finding any interesting patterns or facts to make a relation between these connecting dots mentioned in these research papers.

## 6. EXPLORATORY DATA ANALYSIS

### Impact of COVID-19 on Burning Activities

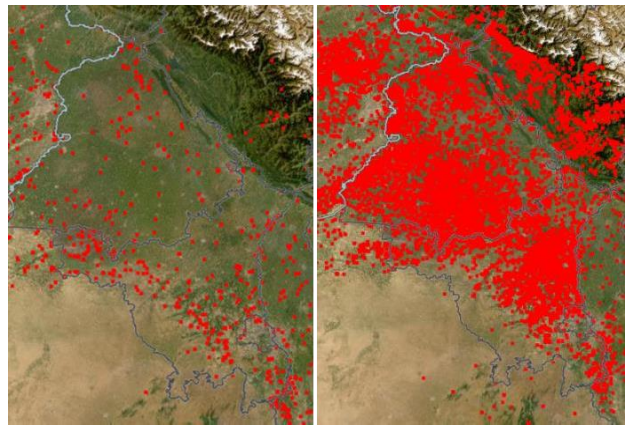

Figure 8 Punjab April 2020    Figure 9 Punjab April 2022 (Source: 31)

A significant difference in the fire activities is visible on the map, majority of the fires will be related to stubble burning during this period. However, with time things come to normal.

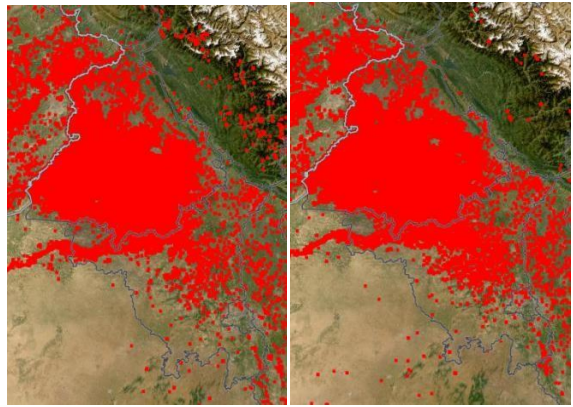

Figure 10 Punjab November 2020　　Figure 11 Punjab November 2022

As a result, stubble burning kept on going as normal during the major peak period of COVID year.

### Heatmap

Further, to verify the concentration of AQI in these three states individually a head map has been created in Tableau.

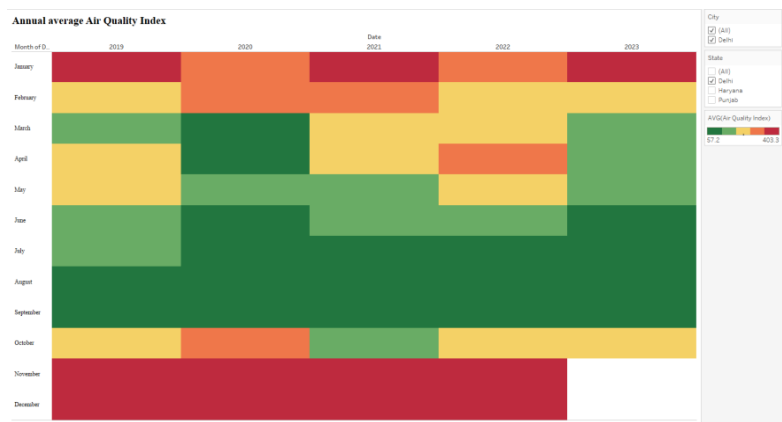

Figure 12 Heatmap of monthly average AQI in Delhi from 2019-2023

It is clear from the heatmap that AQI was at its peak during November and December, even in COVID year. However, during the immediate COVID period, AQI was at its lowest and a similar pattern was noticed in the year 2023.

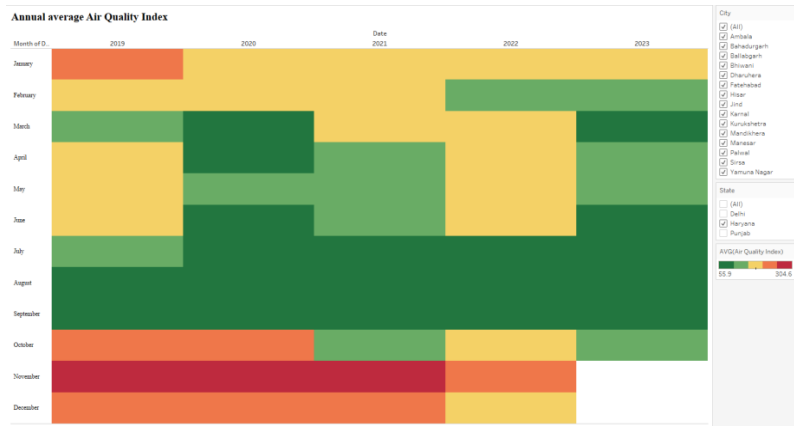

Figure 13 Heatmap of monthly average AQI in Haryana from 2019-2023

As compared to Delhi, in Haryana, the AQI is lower during January, October, and December in all the years. Whereas, the high AQI is visible during the peak period.

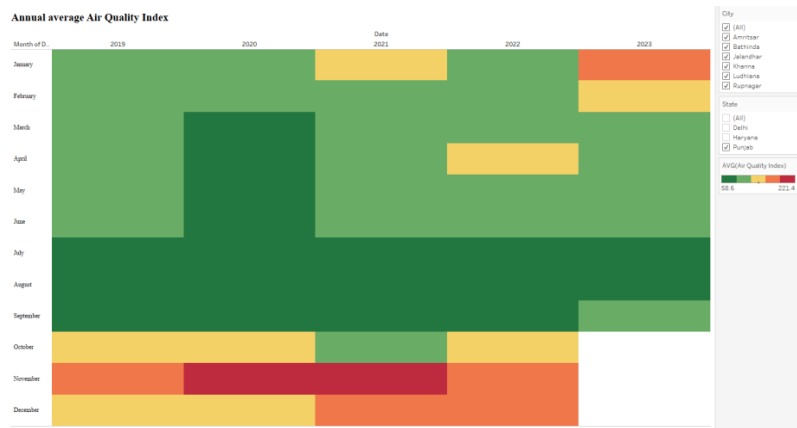

Figure 14 Heatmap of monthly average AQI in Punjab from 2019-2023

As compared to Delhi and Haryana, the Air quality of Punjab is healthier to survive.

**Seasonality and Relation between Different Air Pollutants and AQI.**

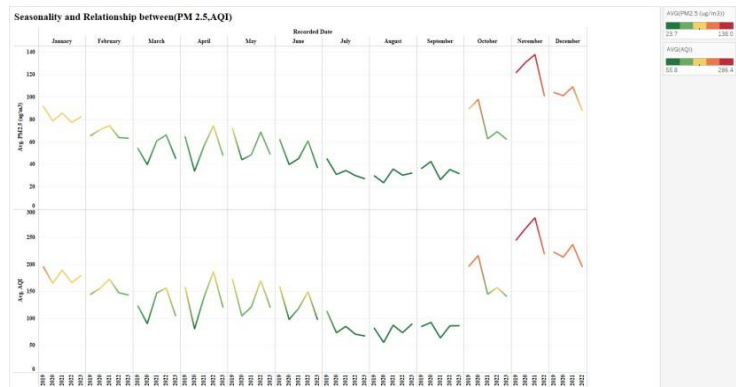

Figure 15 PM 2.5/ AQI

25

The purpose of this graph is to show, the relation between PM 2.5 and AQI, as both the measures change in the same way and at the same time during the year. Whenever the concentration of PM 2.5 increases in the air, the AQI automatically increases the same way and vice versa. Both PM 2.5 and AQI have a peak period of almost three months from October to December each year.

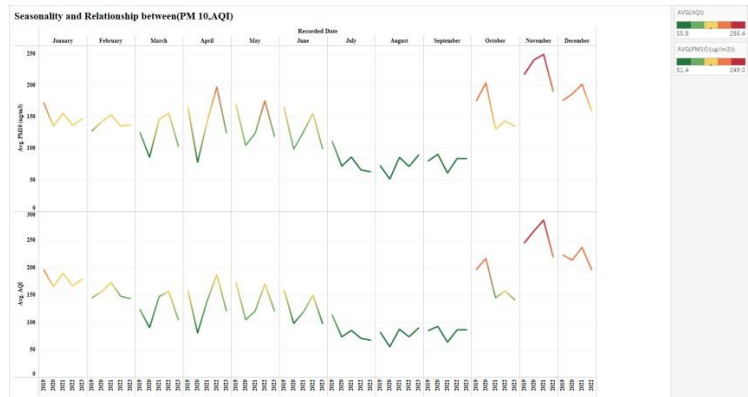

Figure 16 PM10/AQI

The same way it changes for PM 10 as well. Even, the steps of upward or downward trends are the same. Thus, the air pollutants taken into consideration for this research are highly correlated with AQI. So, a control or at least a careful lifestyle should be opted to be healthier in this deadly air.

## 7. DATA SPLIT

As it is a time series data, so the test train data has been split by date rather than percentage or ratio like (80% test, 20% split). However, even based on date it is approximately divided into 4 years for training and approximately one year for testing. For example, from 01-01-2019 to 01-10-2022 is the training data, and the remaining from 01-10-2022 to 15-10-2023 is taken as testing data. AQI is the target variable (dependent) which is predicted based on other air pollutants(Independent) variables.

## 8. MACHINE LEARNING, TIME SERIES AND DEEP LEARNING MODEL

### Machine Learning Models

#### CatBoost

CatBoost is based on the theory of decision trees and gradient boosting. The main purpose of boosting is to sequentially combine many weak models and thus through greedy search create a strong competitive predictive model. In the decision tree growth procedure, CatBoost follows a different approach than gradient boosting models. CatBoost constructs oblivious trees, meaning that all nodes are levelled equally, employing the same predictor and condition for testing. Hence, an index of a leaf can be calculated with bitwise operations.

An irregular organization of gradient learning data is employed to enhance the implementation of the CatBoost algorithm and mitigate the risk of overfitting. The training ability of the CatBoost algorithm is managed by its framework hyperparameters, like iterations number, rate of learning, maximum depth etcetera.

The parameters considered in this research are listed below.
Iterations: The number of steps the algorithm takes to create a more accurate model that learns from the data. Keeping the iterations to a lower number is a good step for optimization.
learning_rate: It scales the contribution of each decision tree to maintain a balance between model accuracy and stability.

depth: It refers to the height of the decision tree. It is used to control over-fitting.
However, in this research 'depth': [3, 6, 8], 'learning_rate': [0.01, 0.05, 0.1], 'iterations': [100, 300, 500], 'loss_function':['RMSE'], and 'random_seed':[42] are considered.

### XGBoost

XGBoost is one of the important types of ensemble learning algorithms in ML approaches [21]. A technique for combining numerous weak classifiers into a single effective one is known as boosting. It is mentioned by [22] that the basic step for the development of XGBoost is applying Gradient Boosting. XGBoost variation of Gradient Boosting outperforms the original in terms of computing efficiency, scalability, and generalization performance. The most important thing to focus on is, data organization while using XGBoost. XGBoost only takes numeric vectors as an input so transforming categorical data into their numerical equivalent is important before using the model. Considering this point the spatial variables like State and city are converted into numerical versions using One-Hot Encoding. Further, feature engineering and data purification need to be applied for best performance.

To enhance the framework performance, the boosting method constructs diverse trees and sequentially refines the predictions by considering residuals from historical trees. When the tree reaches a predetermined number or when the training stage error cannot be further reduced to the expected sequential tree count it results in a halt for further tree development. For symmetric trees, a random subsample without replacement is used for training, enhancing the adaptability of the tree, and forming an improved framework. The same parameters as of CatBoost model are considered for XGBoost.

### Random Forest Regressor

RF is a type of ensemble ML algorithm proposed by Breiman in 2001 [23]. A combination of bagging and random subspace approaches is the foundation of RF. It involves the process of creating an ensemble method that is used to merge these trees into a single forecast based on the bagging of numerous learning trees, where N examples from the training set (D) are used to form bootstrap samples (Db) of n random instances. These bootstrap samples are used to build K different regression trees. The random forest prediction is made by averaging K predictions from regression trees hk (x).

The main parameters for RFR are n_estimators, max_features and max_depth.
n_estimators: The number of trees in the forest.

max_features: The maximum number of features considered for splitting a node.
For this task 'n_estimators': [100, 200, 500], 'max_depth': [3, 6, 8], 'max_features': ['auto', 'sqrt', 'log2'], and 'random_state': [42] is passed for the tuning of RFR.

**Support Vactor Regressor**

Support Vector Regressor is a type of supervised ML model proposed by [24] in 1997. In SVR, the training data consists of predictor variables and observed responses. The main goal of SVM is to find a function f(x) that deviates from yn (sample labels) by a value no greater than bias for each training point x that is, remains as flat as possible. Thus, SVM is also known as Tube regression.

The kernel functions used in SVR, including linear, polynomial, radial basis, and sigmoid functions, facilitate this process. An optimal hyperplane is employed for SVM which minimizes the distance from all data points, effectively approximating the training points and minimizing prediction errors. It is mentioned in [25] that the choice of kernel functions and their parameters significantly affects model accuracy. Methods like grid search using cross-validation approaches are an optimal approach to determine the optimal values for these parameters. Resultantly, it offers the highest cross-validation accuracy, ensuring robust and effective SVR modelling.
There are three main parameters to consider while using the regression model.

Kernal: It is used to transform the data into the required form—for example, linear, nonlinear, polynomial, radial basis function (RBF), and sigmoid. The default is rbf.

Gamma: It provides insight into the extent to which individual data points influence the decision boundary.

C: A hyperplane that maximizes the margin, as it tends to generalize better on unseen data. However, it is important to keep in mind, that setting C too small can lead to underfitting, where the model fails to capture the underlying patterns in the data.

For this model kernel='rbf', C=100, epsilon=0.1 are considered.

**Time Series Model**

**SARIMAX**

SARIMAX model is an abbreviation for Seasonal Autoregressive Integrated Moving Average or Seasonal ARIMA and belongs to the family of the ARIMA model. It is mentioned in [11] that the SARIMAX model is capable of handling time series data that has a single variable along with seasonality. Apart from the hyperparameters for ARIMA, three different hyperparameters namely the AutoRegression (AR), Integrated (I), and Moving Average (MA) for the seasonal part of the series along with one parameter for the period of the seasonality. Just adding X (an exogenous variable) in the SARIMA model makes it SARIMAX.

The SARIMA model comprises two significant elements known as Trend and Seasonal whose hyperparameters are to be selected to configure the SARIMA model.
TREND elements:

p: Trend-based autoregression order.
d: Trend-based difference order.
q: Trend-based moving average order.
SEASONAL elements:
P: Seasonal based autoregressive order.
D: Seasonal-based difference order.
Q: Seasonal-based moving average order.

m: The count of time units for a single season (or a seasonal period).

In a research paper [11] they used SARIMAX to predict NO2 for Faridabad and Delhi city based on six years of records. But at one time it was used to predict just one air pollutant for just one city. In the same way, the SARIMAX model cannot be applied to predict AQI for all these monitoring stations at once. At one time it can predict for one station only. As I am working on a huge data, it is not possible to predict for each monitoring station using SARIMAX. So, I have chosen the most polluted city, which is the capital of India known as Delhi to predict AQI. A separate data frame of Delhi air pollutants concentration during the same time frame has been created from the original dataset used for this research. SARIMAX is used when there is presence of seasonality in the data. So, the dataset of Delhi state has been checked for seasonality.

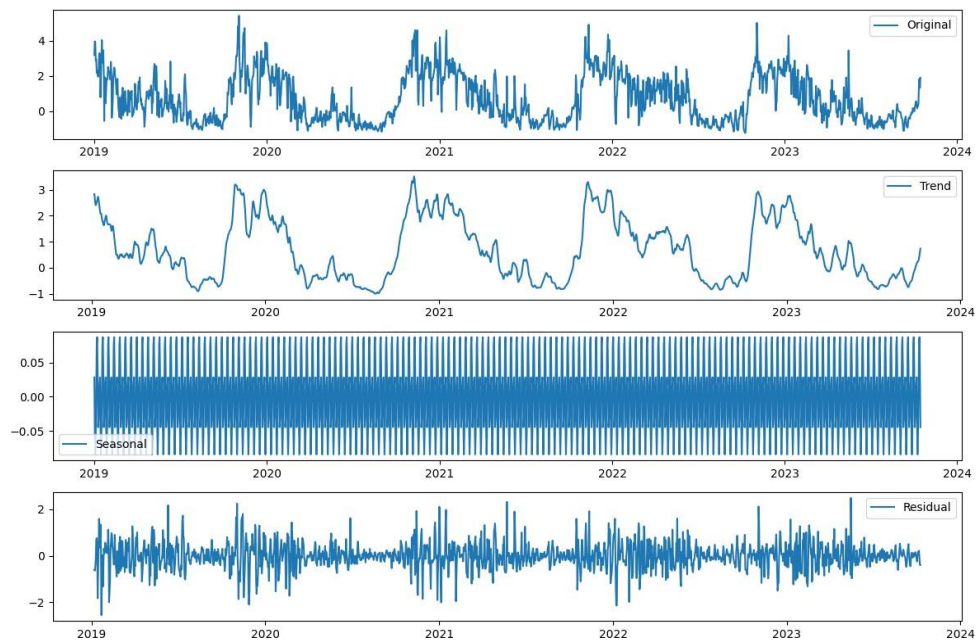

Figure 17 Seasonality components of the Delhi AQI dataset

It can be noticed that there is seasonality in the dataset. The figure reveals that during the fourth quarter of each year, the concentration of AQI was higher as compared to the rest of the time. In other words, Delhi's air pollution peaks during the winter months starting with Diwali (a festival when excessive fireworks are used) and post-harvest agricultural waste burning and deteriorates further with lower surface temperatures increasing in demand for space heating [26] [27] [28].
Before applying the SARIMAX model different tests are used to analyze the dataset and ensure the model is trained accurately.

The Dickey-Fuller test plays an important role to check if the AR model has unit root (null hypothesis) and if the data has seasonality or trend-seasonality (alternate hypothesis), mainly it is used to determine whether the data is stationary or non-stationary.

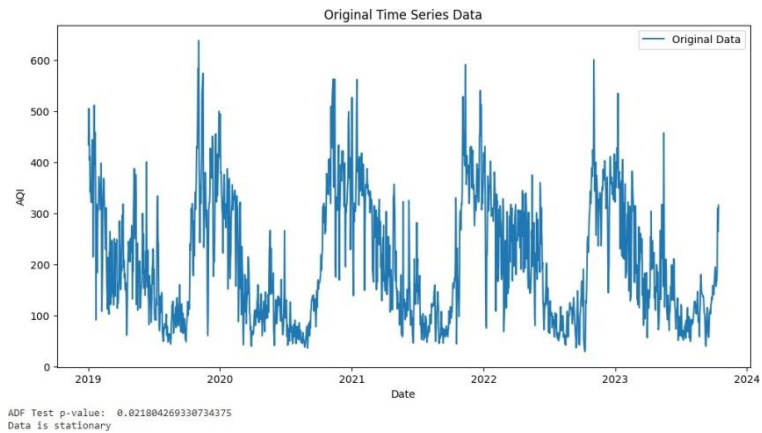

Figure 18 The Dickey-Fuller test

Thus, the test results depict that the data is stationary. So, no further experimentation is needed for the dataset.

To choose the right SARIMAX hyper-parameters the Autocorrelation and partial autocorrelation function plots are created. ACF and PACF help to identify the lags that have high correlations.

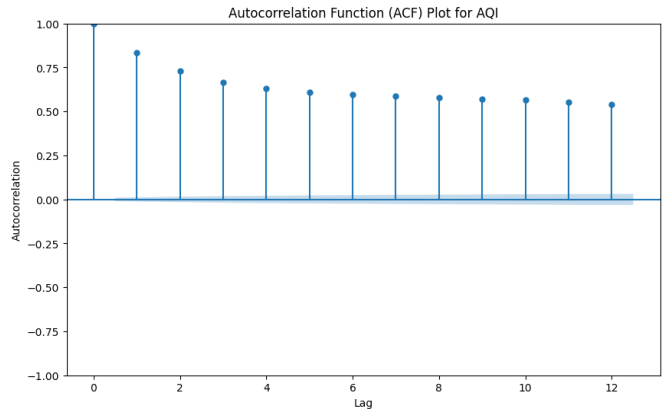

Figure 19 ACF

It is already clear that the time series are seasonal and the ACF plot confirms this pattern. However, by plotting more lags it can be observed if the significance of the lags is gradually declining.

It can be analysed that the highest lag is 1 followed by a gradual decline till 4 and remain constant till 12. It has observed the peak period from the last three to four months October, November, and December. In the same way, it reflected a peak in initial the initial three to four months.

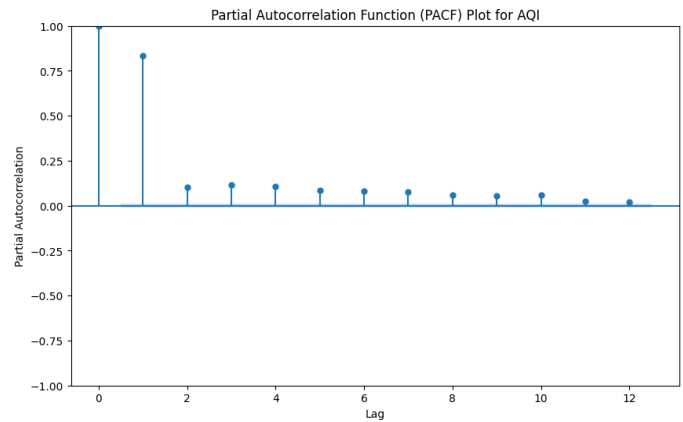

Figure 20 PACF

The lag at 1 is the highest among all others. However, the rest all the last are even below 0.05.

**Deep Learning Model**

**LSTM**

LSTM is a member of Recurrent Neural Networks (RNN) and is distinguished by its capacity to retain memory states over long-term periods, facilitated by its memory cell. This cell encompasses gates responsible for controlling data flow within the memory. The LSTM model comprises three crucial layers: the Input layer, the recurrent layer, and the Output layer. The output layer interfaces with the cell's three gates—input, output, and forget gates—and culminates in producing the final output. The output layer is connected to the three gates of the cell (input, output, and forget gates), and finally, the output layer [29].

After some steps of experimentation, I got the above value as the best performer for LSTM. However, more experimentation can be done to find the optimal hyperparameters.

## 9. RESULTS AND DISCUSSION

During the preprocessing phase, ready-to-use data is created to use for applying different models. The final version of the dataset contains eight air pollutant concentrations PM2.5 (ug/m3), PM10 (ug/m3), NO (ug/m3), NO2 (ug/m3), NOx (ppb), NH3 (ug/m3), SO2 (ug/m3), CO (mg/m3) as an input to predict AQI. Also, it includes One Hot encoded column of three states and 22 cities it has one monitoring station in each city. As it is a time series data, the index is set as a time stamp.

For enhancing the performance of models GridSearchCV is used to find the best parameters for each model. The results of each model are accessed using visualization of the predictions done by each model, performance matrices and residual distribution. The resulting performance metrics are more important than just applying the models. For accurately evaluating the performance metrics different statistical measures are used like MSE, RMSE, R2, and MAE. Different prediction models are used to predict AQI so using these metrics can help to monitor the model's performance.

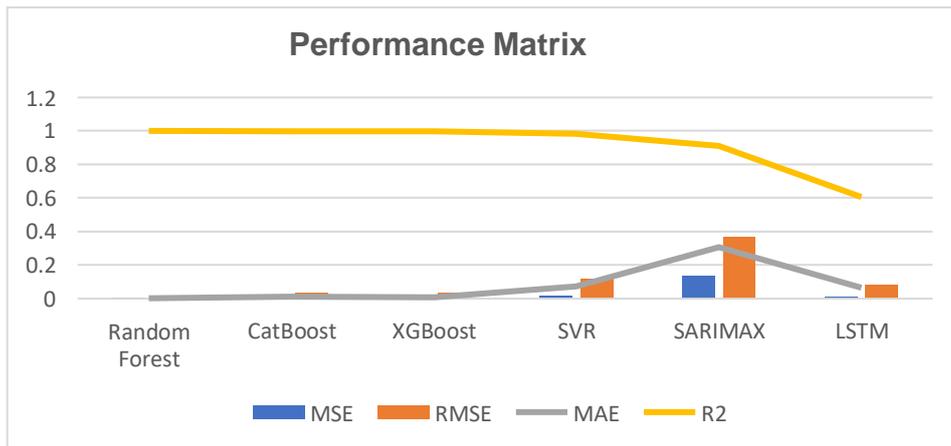

Figure 21. Performance Matrix of all the used Models

The main matrix for the model's performance evaluation especially for regression models is R2. Mainly just four ML models are considered for comparison because SARIMAX has not been used for the whole dataset as other models. So, comparing it with other models is not a good approach. But it can be considered when predicting for same set of locations using all the models. It is clear from figure 21 that the value of R2 is almost 1 for the Random Forest regressor followed by CatBoost and XGBoost ML models. It means these models have explained all the variability in the dependent variable. Also, the other matrices are quite low which indicates, that there are fewer errors made while predicting the actual values. Overall, RFR seems to be the best model among others used in this research. Also, the residuals are tightly scattered around 0.0, which means on average model is predicting correctly. In other words, the model is not systematically overpredicting or underpredicting. However, for the time series and deep learning models, highly computational resources are needed especially for LSTM.

The major limitation of this research is the lack of desired datasets. There is almost no data about meteorological factors like Temperature, humidity, and wind direction etcetera.

## 10. FUTURE SCOPE

ML regression models like Random Forest, CatBoost, and XGBoost have performed well on this dataset, further, these models can be used on similar datasets. It is recommended to use the meteorological data as well if available, as it might come up with new findings. Further, SARIMAX can be used to predict AQI for remaining states like Haryana and Punjab or specific cities. LSTM model can be applied for similar data by following the required preprocessing steps.

## 11. CONCLUSION

Various authors have implemented ML models to solve the same problem that has been handled in this paper. Researchers have implemented models like CatBoost Regressor, SVM, SARIMAX, Random Forest etcetera for various AQI components. In most of the reviewed research papers, preprocessing steps like missing values and outliers are not handled carefully. In this paper, almost all the preprocessing steps are followed as per the needs of the data. The biggest issue that affects any ML model is the existence of null values, which is handled very well in this research using suitable imputation methods. To evaluate the performance of the used models MSE, RMSE, MAE, and R2 metrics are used. The comparison shows that Random Forest performed the best and then closely followed by CatBoost and XGBoost regressor models. Similarly, it is highlighted in the papers on similar topics that ML regressor models especially random forest is good at predictions. It is found that temporal variables are important to consider while using any prediction model, so temporal future trends and insights can be explored. Almost all the reviewed research papers lack meteorological factors which limits the research to explore its effect on the current issue.

## 12. DECLARATION OF COMPETING INTEREST

We do not have any competing interests.

## 13. SUBMISSION DECLARATION AND VERIFICATION

The work described in this project has not been published previously, also it is not under consideration for publication elsewhere.

## 14. USE OF INCLUSIVE LANGUAGE

While doing this project inclusive language guidelines are considered to acknowledge diversity, respect every individual, and promote equal opportunities. To ensure Gender Neutrality, by default plural nouns are used to avoid gender-specific pronouns. The content is free from bias, stereotypes, and assumptions about personal attributes. To eliminate unintentional biases, the use of language is assessed regularly. By following inclusive language guidelines, this report aims to communicate effectively while embodying a commitment to inclusivity.